%% file: IEEE-conference-template-062824.tex
\documentclass[conference]{IEEEtran}
\IEEEoverridecommandlockouts
\usepackage{amsmath,amssymb,amsfonts,amsthm}
\newtheorem{theorem}{Theorem}[section]

\theoremstyle{definition}
\newtheorem{definition}[theorem]{Definition}

\theoremstyle{remark}

\usepackage[square,numbers]{natbib}
\usepackage{algorithmic}
\usepackage{graphicx}
\usepackage{textcomp}
\usepackage{xcolor}
\usepackage[utf8]{inputenc} % allow utf-8 input
\usepackage[T1]{fontenc}    % use 8-bit T1 fonts
\usepackage{hyperref}       % hyperlinks
\usepackage{url}            % simple URL typesetting
\usepackage{booktabs}       % professional-quality tables
\usepackage{nicefrac}       % compact symbols for 1/2, etc.
\usepackage{microtype}      % microtypography
\usepackage{arydshln}
\usepackage{soul}
\DeclareMathOperator*{\argmax}{arg\,max}

\usepackage{subcaption}

\usepackage[textsize=tiny]{todonotes}

\usepackage[capitalize,noabbrev]{cleveref}
\usepackage{enumitem}

\crefformat{section}{\S#2\color{blue}#1#3} % see manual of cleveref, section 8.2.1
\crefformat{subsection}{\S#2\color{blue}#1#3}
\crefformat{subsubsection}{\S#2#1#3}

\definecolor{mygray}{gray}{0.95}

\usepackage{framed}
\usepackage{fancybox}
\colorlet{shadecolor}{mygray}
\colorlet{framecolor}{white}

\usepackage{tcolorbox}

{\endMakeFramed}

\newenvironment{froval}{%
\MakeFramed {\advance\hsize-\width\FrameRestore}}%
{\endMakeFramed}

\newcounter{cfinding}
\newenvironment{cfinding}[1][]{\refstepcounter{cfinding}\par\medskip
   \noindent\textbf{Position~\thecfinding. #1} \rmfamily}{\medskip}

\usepackage{xspace}
\newcommand{\adaptiverob}{continual adaptive robustness\xspace}

\begin{document}

\title{Position: Towards Resilience Against Adversarial Examples\\
}

\author{\IEEEauthorblockN{Sihui Dai}
\IEEEauthorblockA{\textit{Princeton University}\\
sihuid@princeton.edu}
\and
\IEEEauthorblockN{Chong Xiang}
\IEEEauthorblockA{\textit{Princeton University}\\
cxiang@princeton.edu}
\and
\IEEEauthorblockN{Tong Wu}
\IEEEauthorblockA{\textit{Princeton University}\\
tongwu@princeton.edu}
\and
\IEEEauthorblockN{Prateek Mittal}
\IEEEauthorblockA{\textit{Princeton University}\\
pmittal@princeton.edu}
}
%\and
%\and

\maketitle

\begin{abstract}
\input{sections/1abstract}

\end{abstract}


\section{Introduction}
\input{sections/2intro}
\section{Background: Adversarial Robustness}
\input{sections/3current_AT}

\section{Towards Resilience Against Adversarial Examples}
\input{sections/4position}

\section{Continual Adaptive Robustness: A Simplified Setting of Adversarial Resilience}
\input{sections/7CAR}

\section{Connection to Simultaneous Multiattack and Unforeseen Attack Robustness}
\label{sec:connection}
\input{sections/5MAR}

%\section{Unforeseen Attack Robustness}
%\input{satml2024/sections/6UAR}

\section{Conclusion}
\input{sections/8conclusion}

\bibliography{example_paper}

%%%%%%%%%%%%%%%%%%%%%%%%%%%%%%%%%%%%%%%%%%%%%%%%%%%%%%%%%%%%%%%%%%%%%%%%%%%%%%%
%%%%%%%%%%%%%%%%%%%%%%%%%%%%%%%%%%%%%%%%%%%%%%%%%%%%%%%%%%%%%%%%%%%%%%%%%%%%%%%
% APPENDIX
%%%%%%%%%%%%%%%%%%%%%%%%%%%%%%%%%%%%%%%%%%%%%%%%%%%%%%%%%%%%%%%%%%%%%%%%%%%%%%%
%%%%%%%%%%%%%%%%%%%%%%%%%%%%%%%%%%%%%%%%%%%%%%%%%%%%%%%%%%%%%%%%%%%%%%%%%%%%%%%
\newpage
\appendix
\input{sections/appendix}

\end{document}

%% file: sections/1abstract.tex
%Robustness against adversarial examples is an important field in machine learning research that looks at improving the reliability of models in the presence of a test-time adversary. 
%\prateek{the following sentence is a bit awkward}
%In safety-critical applications such as autonomous driving, adversarial robustness is crucial for safe deployment.
Current research on defending against adversarial examples focuses primarily on achieving robustness against a single attack type such as $\ell_2$ or $\ell_{\infty}$-bounded attacks. However, the space of possible perturbations is much larger than considered by many existing defenses and is difficult to mathematically model, so the attacker can easily bypass the defense by using a type of attack that is not covered by the defense. In this position paper, we argue that in addition to robustness, we should also aim to develop defense algorithms that are \textit{adversarially resilient} --- defense algorithms should specify a means to quickly adapt the defended model to be robust against new attacks.  We provide a definition of adversarial resilience and outline considerations of designing an adversarially resilient defense.  We then introduce a subproblem of adversarial resilience which we call \textit{continual adaptive robustness}, in which the defender gains knowledge of the formulation of possible perturbation spaces over time and can then update their model based on this information.  Additionally, we demonstrate the connection between continual adaptive robustness and previously studied problems of \textit{multiattack robustness} and \textit{unforeseen attack robustness} and outline open directions within these fields which can contribute to improving continual adaptive robustness and adversarial resilience.
 %The discrepancy between the focus of current defenses and the space of attacks of interest calls to question the practicality of existing defenses and the reliability of their evaluation.  
%In this position paper, we argue that the research community should look beyond single attack robustness, and we draw attention to three potential directions involving robustness against multiple attacks: \textit{simultaneous multiattack robustness}, \textit{unforeseen attack robustness}, and a newly defined problem setting which we call \textit{\adaptiverob}.
%We provide a unified framework which rigorously defines these problem settings, synthesize existing research in these fields, and outline open directions.  We hope that our position paper inspires more research in simultaneous multiattack, unforeseen attack, and \adaptiverob.

\begin{figure*}[hbtp]
    \centering
    \includegraphics[width=\textwidth]{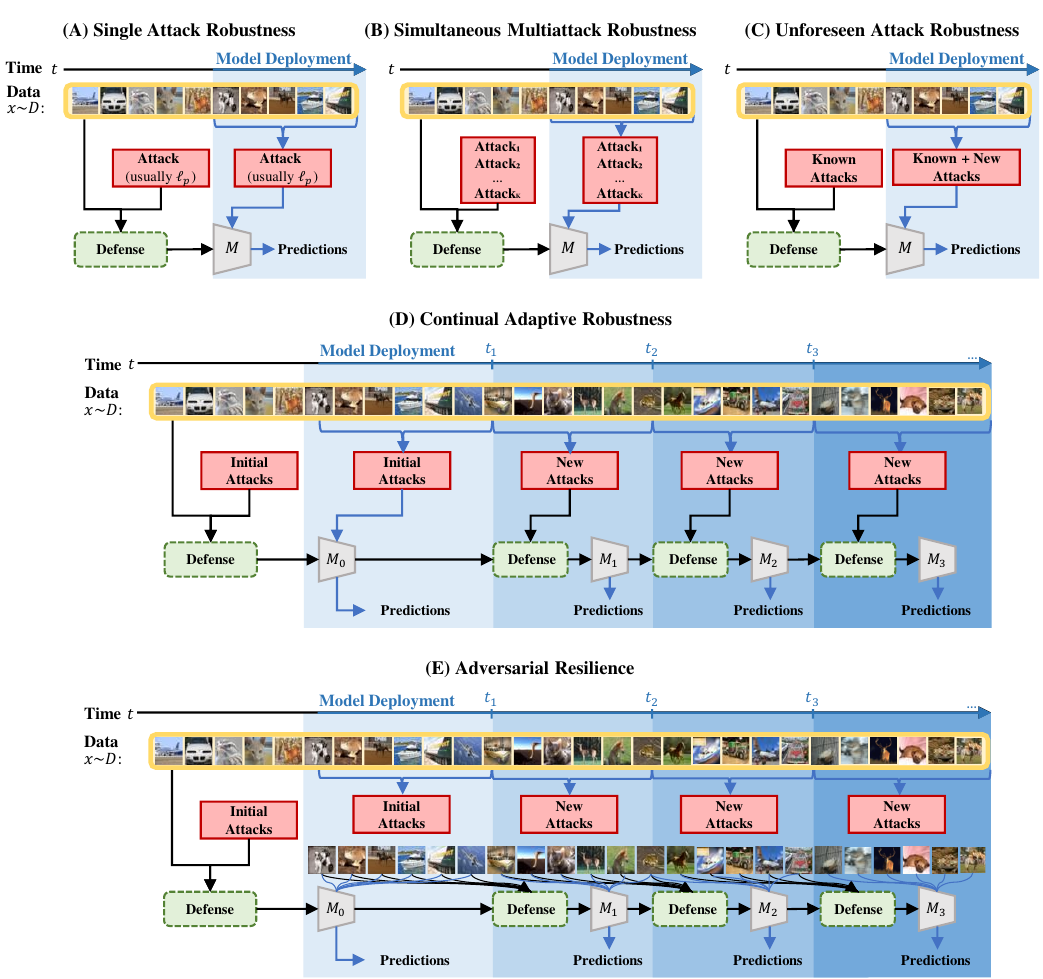}
    \vspace{-15pt}
    \caption{\textbf{\underline{Overview of directions in adversarial robustness research}}: \textbf{(A) Single Attack Robustness (\S\ref{Sec:background}).}  This is the setting commonly studied in prior work.  The defender uses information about a single attack (usually an $\ell_p$-bounded attack) to output a robust model.  This robust model is deployed, and an attacker attacks the model using the same attack type. \textbf{(B) Multiattack robustness (\S\ref{Sec:background})}.  The defender aims to obtain robustness against a set of $K$ different attacks simultaneously.  The defender can use information about these attacks to output a robust model. After deployment, an attacker attacks the model with the same set of $K$ attacks. \textbf{(C) Unforeseen attack robustness (\S\ref{Sec:background}).}  After deployment, the attacker attacks the defended model with attacks that were not considered in the design of the defense. \textbf{(D) Continual adaptive robustness (\S\ref{Sec:CAR}).} \textit{New setting introduced in this paper.} Over time, the adversary can introduce new attacks, and similarily the defender can use information about the previous model and details of the new attack type to update the deployed model to improve robustness against these new attacks. \textbf{(E) Adversarial resilience. }\textit{New setting introduced in this paper.}  Similar to (D), the adversary can introduce new attacks, but now the defender only has access to perturbed examples seen during test-time (instead of details of new attack algorithms) in order to update the defended model.
    }
    \label{fig:overview}
    \vspace{-10pt}
\end{figure*}

%% file: sections/2intro.tex
While current machine learning (ML) models are able to achieve high accuracy on classification tasks such as image classification, these models misclassify when a small imperceptible perturbation is added to the input during test time.  This perturbed image is known as an adversarial example \citep{szegedy2013intriguing}.  These adversarial examples can be exploited by an adversary to cause harm in safety-critical applications and raise doubts about model reliability.

Adversarial robustness research consists of two main components: researching attacks in order to better evaluate robustness and designing stronger defenses that can achieve good performance under the presence of these attacks.  In the field of computer vision, research in designing attacks has led to a wide variety of attacks following different threat models, including spatial transformations \citep{XiaoZ0HLS18, engstrom2019exploring}, color shifts \citep{LaidlawF19}, JPEG-compression based attacks \citep{kang2019robustness}, weather-based attacks \citep{Schmalfuss_2023_ICCV, kang2019robustness}, Wasserstein distance bounded attacks \citep{wasserstein_attacks, wu2020stronger}, and perceptual distance based attacks \citep{laidlaw2020perceptual, ghazanfari2023r}.  However, most existing work on defenses has focused on defending against specific narrow threat models (primarily $\ell_{\infty}$ and $\ell_2$ bounded adversaries) \citep{madry2017towards, zhang2019theoretically, cohen2019certified, croce2020robustbench}. This discrepancy between the focus of current defenses and the space of existing attacks leads to vulnerability; an attacker can easily breach the defense by using an attack different from the focus of the defense.  

A few works have looked at training with multiple attacks \citep{TB19, MainiWK20, madaan2020learning, croce2022adversarial} or improving generalization to unforeseen attacks \citep{laidlaw2020perceptual, dai2022formulating}.  Figure \ref{fig:overview} provides a diagram of these problems and how they compare to the commonly studied setting of single attack robustness. 
 However, existing defenses in these settings cannot guarantee satisfactory performance; \citet{dai2023multirobustbench} demonstrate that when evaluating these techniques on a wide space of attacks (including unforeseen attacks), the robustness of the best-performing approach on the worst-case attack is 3.30\%.

The current progress in adversarial robustness makes the future of this field seem quite bleak; given that new types of attacks can be discovered over time, it seems unlikely that a single model can achieve reasonable robustness against all current and future attacks types.  We argue that the main limiting factor is the definition of robustness itself; robustness is a game between the defender and attacker where the defender defends the model and deploys it and then the performance of the model is measured via an attacker at test-time \cite{huang2011adversarial} (See Figure \ref{fig:overview}ABC for a visualization of previously studied robustness problems).  However, post-deployment the defender may still learn information about new attacks and should update their deployed model as quickly as possible to perform well against the new attacks.  Ability to update and recover from disruptions (such as those caused by new types of attacks) is a security property known as resilience \citep{firesmith_2019}.  In this position paper, we should design robust defenses that also provide way of updating the model post-deployment in order to achieve resilience.  We focus on recovering from disruptions caused by new types of perturbations that were not anticipated during the model development stage and call this property \textit{adversarial resilience}.

The rest of this paper is organized as follows, we first provide background on existing research in adversarial robustness and provide the definition of adversarial robustness provided in \citep{huang2011adversarial} in Section \ref{Sec:background} and discuss limitations of existing robustness works and the robustness definition.  We then present our main position which is that we, as a research community, should develop adversarially resilient defenses, provide a definition for adversarial resilience in Section \ref{Sec:position}, outline possible applications in which adversarial resilience is important in Section \ref{sec:applications}, and discuss challenges of implementing adversarially resilient defenses in Section \ref{sec:ars_challenges}.  We then introduce a simplified problem setting for resilience which we call \textit{\adaptiverob} (CAR) and outline directions for research in CAR in Section \ref{Sec:CAR}.  Finally, in Section \label{sec:connection}, we demonstrate the relationship between \adaptiverob to multiattack robustness and unforeseen attack robustness research \citep{dai2023multirobustbench}.

%% file: sections/3current_AT.tex
\label{Sec:background}
In this section, we introduce the game formulation of adversarial robustness to define the adversarial robustness problem (\S\ref{sec:game}) and  then discuss attacks and defenses proposed in prior works in \S\ref{sec:attacks} and \S\ref{sec:defenses}. 

\begin{table*}[]
    \centering
    \begin{tabular}{c|l|l}
        Notation & Description & Example \\ \hline
        $\mathcal{D}$& data distribution ($\mathcal{D} = X \times Y$) & CIFAR-10 images\\
        $\mathcal{H}$ & space of defended models & space of NNs \\
        $\mathcal{A}$ & defense algorithm, outputs robust model $M$ & adversarial training \citep{madry2017towards, zhang2019theoretically} \\
        $T$ & attack space ($T: X\to 2^X$, space of possible perturbations for each input) &  $\ell_2$ ball of radius 0.5: $T(x) = \{x' | ~ ||x' - x||_2 \le 0.5\}$ \\
        $P_T(x, y, M)$ & worst case perturbation within attack space $T$ (See Eq. \ref{eq:atk_obj}) & \\
        $\tilde{P}_T(x, y, M)$ & attack algorithm for searching for an attack in $T$ & $\ell_2$ PGD  \citep{madry2017towards} \\
        $K$ & set of attack spaces $T$ & $K = [\{x' | ~ ||x' - x||_2 \le 0.5\}, \{x' | ~ ||x' - x||_\infty \le \frac{8}{255}\}]$
    \end{tabular}
    \caption{Notations used across paper}
    \label{tab:my_label}
\end{table*}

\subsection{A Game Formulation of Single Attack Robustness}
\label{sec:game}
A useful notion for defining adversarial robustness is an adversarial game between a defender and an attacker \citep{huang2011adversarial,bruckner2012static,bulo2016randomized,pinot2020randomization,pydi2021many,meunier2021mixed,balcan2023nash}.  This game defines robustness (which is the goal of the defender) by specifying the capabilities and limits of the attacker and defender and the criteria for the defender to win this game.  We will now provide a game formulation (adapted from \citep{huang2011adversarial}) of the setting commonly studied in prior works, which we call \textit{single attack robustness}.

We will use $\mathcal{D} = X \times Y$ to denote the data distribution and $\mathcal{H}$ to denote the space of defended models.  

\begin{definition}[Single Attack Game] \label{def:adv_game}
\end{definition}
\begin{enumerate}[leftmargin=*]
    \item The defender chooses a defense algorithm $\mathcal{A}_T: D \rightarrow \mathcal{H}$ for obtaining a robust model using knowledge of the attack space $T: X\to 2^X$ (space of possible attacks for each input).
    \item   The attacker then chooses an attack procedure $\tilde{P}_T: X \times Y \times \mathcal{H} \to X$ to apply after deployment which outputs a perturbation within $T$ for each input.  The attacker may also use information about the defense algorithm $\mathcal{A}_T$ to determine the attack procedure.
    \item Before deployment, the defender uses the defense algorithm to obtain a robust model: $M = \mathcal{A}_T(D_{\text{train}})$. 
    \item After deployment, for every test data $(x, y) \sim \mathcal{D}$, the defender evaluates the performance of $M$ on the perturbed input $\tilde{P}_T(x, y, M)$.  This performance is assessed using a loss function $\ell: Y \times Y \to \mathbb{R}$.
    \item The defender wins if with high probability, $M$ achieves robust error less than a threshold $\delta_r$ which can be tolerated for the task ($\mathbb{E}_{(x, y) \sim \mathcal{D}} [\ell(h(\tilde{P}_T(x, y, h)), y)] \le \delta_r$) and achieves clean error less than a threshold $\delta_c$ which can be tolerated for the task ($\mathbb{E}_{(x, y) \sim \mathcal{D}} [\ell(h(x)), y)] \le \delta_c$).
\end{enumerate}   
From the above game formulation of single attack robustness, we can see that the goal of the defender is to achieve small loss while the attacker aims to prevent this.  Meanwhile, the constraints on the defender and adversary is that they can only use attacks in a single attack space $T$ and the capability of the defender is the choice of $\mathcal{A}$, while the capability of the attacker is the choice of $\tilde{P}$.  We will now discuss these constraints and capabilities in further detail and discuss prior works. 

\subsection{Attacks in Adversarial ML Research}
\label{sec:attacks}

In the second step of the game formulation of adversarial ML (Definition \ref{def:adv_game}), the attacker has flexibility in choosing the attack procedure while conforming to a set attack space.  We will now discuss different aspects of the attack procedure in more detail.

\textbf{Attack spaces. } Attack spaces ($T:X \to 2^X$) model the space of perturbations that the adversary can make on the input.  For example, a possible attack space is the space of $\ell_2$-bounded perturbations with radius 0.5, which can be modeled via $T(x) = \{x' | ~ ||x' - x||_2 \le 0.5\}$. The attack space can be considered parameterized by an attack type (e.g., $\ell_2$ attack) and an attack strength (e.g., 0.5). 

In the direction of attack spaces, several works look at different forms of perturbations that can occur in practice. One of the most commonly used attack spaces across attack and defense literature is $\ell_p$-bounded attacks \citep{szegedy2013intriguing,madry2017towards,carlini2017towards}.  Outside of $\ell_p$ attacks, a wide array of imperceptible attacks have been introduced, such as imperceptible spatial transformations \citep{XiaoZ0HLS18}, color shifts \citep{laidlaw2020perceptual}, JPEG-compression based attacks \citep{kang2019robustness}, Wasserstein distance bounded attacks \citep{wasserstein_attacks, wu2020stronger}, and LPIPS distance bounded attacks \citep{laidlaw2020perceptual, ghazanfari2023r}.  For computer vision applications models robustness against small distortions is also important.  Along this line, several works have proposed looking at weather-based attacks \citep{kang2019robustness, Schmalfuss_2023_ICCV}, small spatial transformations \citep{engstrom2019exploring}, and patch attacks which allow unbounded perturbations only within a small region of the image \citep{brown2017adversarial, eykholt2018robust}.

\textbf{Attack objective and optimization procedures. } The objective of the attacker is to find the worst case perturbation over the attack space.  This can be formulated as: 
\begin{equation}
    P(x, y, M) = \argmax_{x' \in T(x)} \ell(M(x'), y)
    \label{eq:atk_obj}
\end{equation}
Equation \ref{eq:atk_obj} can also be considered the ideal attack and will use the term "attack" to refer to this ideal formulation.

In practice, obtaining the ideal attack is infeasible so an optimization procedure is used to search over this space in order to obtain an adversarial example. We use $\tilde{P}(x, y, M)$, the \textit{attack procedure}, to denote the approximate worst-case perturbation obtained via applying the optimization procedure. An example of an attack procedure $\tilde{P}$ would be to use projected gradient descent (PGD) \citep{madry2017towards} optimization in order approximate $P$ in Equation \ref{eq:atk_obj}.

There have also been many works focused on creating stronger or more efficient attack procedures in order to obtain a more reliable and efficient assessment of robustness in addition to more realistic attacks that only use information obtained from querying the model (black-box attacks) instead of relying on taking model gradients (white-box attacks).  For example, many different methods of generating $\ell_{\infty}$ or $\ell_2$ attacks have been proposed \citep{goodfellow2014explaining, madry2017towards, carlini2017towards, moosavi2016deepfool, croce2020reliable, gowal2019alternative, andriushchenko2020square}.  There has also been research outside of $\ell_p$ attacks: for example, \citet{wu2020stronger} improves optimization procedure for Wasserstein attacks over \citet{wasserstein_attacks}.

\subsection{Defenses in Adversarial ML Research}
\label{sec:defenses}
In this section, we discuss prior works in defenses for defending against single attacks, multiple (known) attacks, and unforeseen attacks. In Appendix \ref{app:instantiating}, we demonstrate how different defenses can be captured within the adversarial game framework (Definition \ref{def:adv_game}).

\textbf{Single Attack Defenses.} Many existing works study the problem of robustness against a single attack type, primarily $\ell_2$ or $\ell_{\infty}$ bounded attacks at a fixed radius. 
 This setting is depicted in Figure \ref{fig:overview}A and assumes that there is only 1 attack during test-time and the defender can use knowledge of what attack it is in order to generate the robust model.  Currently, many defenses are either based on adversarial training \citep{croce2020robustbench} or are certified defenses \citep{cohen2019certified, wong2018provable}.   In adversarial training \citep{madry2017towards, zhang2019theoretically}, the robust model is obtained by training using adversarial examples following the attack space of interest.  provides a benchmark which ranks adversarial training based defenses against on robustness against $\ell_p$ attacks.  Meanwhile certified defenses include randomized soothing \citep{cohen2019certified, zhang2020black, yang2020randomized, salman2020denoised} and bound propagation \citep{wong2018provable, mirman2018differentiable}.  These certified defenses utilize information about the geometry of the attack space in order to give a guarantee of robustness. 
 Outside of $\ell_p$ attacks, there are also certified defenses for localized patch attacks~\citep{chiang2020certified,xiang2021patchguard,xiang2022patchcleanser}, geometric transformations~\citep{yang2023provable}, and natural out-of-distribution data~\citep{lin2020dual}.

\textbf{Defending against multiple attacks. }A major limitation of single attack defenses is that given the large space of existing attacks, an adversary can easily bypass these defenses by using a different attack than considered by the defender.  Outside of single attack defenses, there has also been a small body of works addressing robustness against multiple attacks.  These works can generally be divided into two categories: 
\begin{enumerate}[leftmargin=*]
    \item \textit{Simultaneous multiattack robustness (sMAR)}: sMAR works assume that the defender knows about all attacks it would like to be robust to \textit{a priori} and is thus able to use information about these attacks in order to generate the defended model.  This setting is depicted in Figure \ref{fig:overview}B.  Many works in this direction study how to best incorporate multiple attacks in training \citep{TB19, MainiWK20, madaan2020learning,croce2022adversarial}.  Outside of works on training with multiple attacks \citet{Croce020} propose a certified defense but is restricted to unions of $\ell_p$ balls.
    \item \textit{Unforeseen attack robustness (UAR)}: UAR works to achieve better generalization to a wide variety of attacks without any knowledge of these attacks in advance.  This is depicted in Figure \ref{fig:overview}C. 
 Existing works in this direction generally try to approximate model the space of imperceptible attacks \citep{laidlaw2020perceptual} or use regularization \citep{dai2022formulating, jin2020manifold} in order to improve generalization across attacks.
\end{enumerate}
Additional details about prior works for sMAR and UAR are available in Appendix \ref{app:sMAR} and \ref{app:UAR} respectively. While these directions seem to address the limitations of single attack robustness, \citet{dai2023multirobustbench} demonstrates that when evaluated on a large scope of potential existing attacks, these defenses cannot give satisfactory robustness in the worst case; currently, the best-performing approach on the benchmark achieves on 3.3\% robustness for the worst-case attack in the evaluation set.

\textbf{Limitations of robustness. }We argue that we need to reconsider our goals in research in adversarial robustness.  Given the difficulty of achieving robustness against multiple attacks as we cannot anticipate all the different types of attacks that will exist in the future, we should also ask ourselves, what should we do if we have deployed a model and sometime in the future someone discovers a new attack that our model is vulnerable to? The game formulation of adversarial robustness in Definition \ref{def:adv_game} does not shed any insight in this direction; the defender would lose the game if the attacker successfully increases the error past the allowed threshold $\delta_r$, and thus the next steps would just be to restart the game and choose a different defense algorithm.  However, this formulation neglects goals such as \textit{quick} recovery after the discovery of a successful attack.  Because of this, we argue that we should develop robust defenses that also provide a method of quickly updating the deployed model when new attacks arise.
%\textbf{Limitations. } We argue that achieving single attack robustness gives a false sense of security. Currently, with the wide variety of proposed attack types, the adversary can utilize any of these attack types to attack the model.  This makes the attacker's single attack space constraint in the single attack game (Definition \ref{def:adv_game}) unrealistic.

%Additionally, the paradigm of evaluating a defense only on the single attack that the defense is targeted towards gives an unreliable measure of true robustness.  Currently, defenses against single attacks do not necessarily generalize to other attack types.  For example, \citet{dai2022formulating} demonstrates that a model adversarially trained for $\ell_2$ attacks with radius 0.5 can achieve 66.65\% robust accuracy on those attacks, but only 0.76\% robust accuracy on spatial attacks \cite{XiaoZ0HLS18}.

%The impracticality of research in single attack robustness have also been highlighted by \citet{gilmer2018motivating} and \citet{tramerpre}.  We deviate from these works by specifically highlighting the importance of research in robustness against multiple attacks, synthesizing existing work in this direction, and outlining open problems.

%% file: sections/4position.tex
\label{Sec:position}

To begin, we state our main argument below:

\begin{froval}
    \begin{cfinding}\label{cfind: justification}We urge the research community to study \textit{adversarial resilience} and develop defense algorithms that provide a method of quickly updating the model to be robust against new attacks.
    \end{cfinding}
\end{froval}

We will now give a more formal description of the problem setting of interest and then define \textit{adversarial resilience}.

\textbf{Problem setting for adversarial resilience. }We consider the setting where during test-time, the attacker is actively designing new attack types in order to try to fool the model.  We can model this by considering the space of attacks used by the adversary as a function of time $t$.  Specifically, we model this by what we call an adaptive knowledge set.

\begin{definition}[Adaptive Attacker Knowledge Set] The attacker knowledge set $\mathbf{K}_{\text{attacker}}(t)$ is a time dependent function that outputs a set of attack spaces $T: X \to 2^X$ that the attacker knows about at time $t$, and can thus use attacks algorithms which search over one or more of the attack spaces in this set to attack the model at time $t$.  We note that this function satisfies the property that $\forall t_1 < t_2$, $\mathbf{K}_{\text{attacker}}(t_1) \subseteq \mathbf{K}_{\text{attacker}}(t_2)$ (i.e., once the attacker has discovered a new attack space, they know about this attack space at all later times and can use this when generating new attacks).
\end{definition}

Given the integration of time into the problem setting, we can also consider the defense algorithm itself depending on time.  We now provide a definition of what we call an adaptive defense algorithm.

\begin{definition}[Adaptive Defense Algorithm]
    An adaptive defense algorithm $\mathcal{A}$ is a time dependent learning algorithm which takes in an initial training dataset $D_{\text{train}}$, an initial set of attacks known to the defender $K_{\text{defender}}$, and the adversarial examples generated by the adversary at time $< t$ and outputs a model $M_t$ to be applied at time $t$.
\end{definition}

This formulation of an adaptive defense algorithm allows the defender to update their defense based on previous adversarial examples present during testing, which is a quality that defenses for adversarial robustness currently do not consider. This ability would help the defender in the case where the test-time attacker formulates a new attack type that that the defender is vulnerable to, as the defender would thus be able to update their defense to recover quickly from this attack.  We now introduce the notion of adversarial resilience.

\begin{definition}[Adversarial Resilience] \label{def:resilience}
    Let $L(M, K)$ denote robust error over $\mathcal{D}$ for model $M$ measured over a set of attack spaces $K=\{T_1, ..., T_n\}$ and let $\mathbf{K}_\text{attacker}$ denote the adaptive attacker knowledge set.  Let $t'$ denote a time at which $\mathbf{K}_\text{attacker}$ expands to include a new attack (ie. $\mathbf{K}_\text{attacker}(t) \subset \mathbf{K}_\text{attacker}(t')$ for $0 < t < t'$.  An adaptive defense algorithm $\mathcal{A}$ has \textit{adversarial resilience} if it is able to maintain the following properties:
    \begin{itemize}[leftmargin=*]
        \item \textbf{Good robustness against previously seen attacks: }The defended model $M_{t'}$ at time $t'$ should be robust against attacks conforming to attack spaces $T \in \mathbf{K}_\text{attacker}(t)$ for $t < t'$: $$L(M_{t'},\mathbf{K}_\text{attacker}(t)) \le \delta_k$$
        where $\delta_k > 0$ denotes the amount of error that we can tolerate for attacks previously present in training.
        \item \textbf{Some unforeseen robustness against new attacks: } The defended model $M_{t'}$ at time $t'$ is able to achieve some degree of unforeseen robustness against the new attack(s) added at time $t'$: $$L(M_{t'}, \mathbf{K}_\text{attacker}(t')) \le \delta_u$$
        where $0 < \delta_k \le \delta_u$ represents the amount of degradation that can be tolerated against new attacks.  We note that it is much more difficult to achieve robustness against unforeseen attacks so the amount of error tolerated is larger than for previously seen attacks.
        \item \textbf{Quick recovery from the new attack: }  $\mathcal{A}$ outputs a new model within $\Delta t$ from the introduction of the new attack at time $t'$, where $\Delta t$ is a small value representing how long the algorithm is allowed to take.  This new model at time $\Delta t$ achieves at most $\delta_k$ error when tested on attacks in $K(t')$:
        $$L(M_{t'+\Delta t}, \mathbf{K}_\text{attacker}(t')) \le \delta_k$$
        \item \textbf{Maintains accuracy on unperturbed inputs: }All defended models outputted by $\mathcal{A}$ achieve high clean accuracy:
        $$\forall t ~ \mathbb{E}_{(x, y) \sim \mathcal{D}} \ell(h_t(x), y) \le \delta_c$$
        where $\delta_c$ is the threshold of error on unperturbed inputs that can be tolerated for the application.
    \end{itemize}
\end{definition}
As we can see from Definition \ref{def:resilience}, adversarial robustness can be considered adversarial resilience at a specific time.  The added dimension of time also helps introduce more complexity into the problem of adversarial robustness by allowing for 2 different robustness thresholds $\delta_k$ and $\delta_u$ based on how recently a new attack was introduced.  We note that $\delta_k \le \delta_u$ because in general we should expect less error when we have seen examples of the attack previously than if we have not.  Additionally, the time dimension also allows us to add the requirement that the model recovers quickly from the attack, specifically it must recover to have at most $\delta_k$ robust error within a small time window of $\Delta t$ after the introduction of a new attack.

\textbf{Choices for robust error $L$. }We note that in Definition \ref{def:resilience}, performance is measured with respect to a loss $L$, which takes in a model $M$ and set of attack spaces $K$ that we want to evaluate. $L$ generally provides a way of computing a robust error across a set of attacks using a loss function for single attacks $\ell$.  We now provide some examples of $L$.% \sophie{clarify notation here $P$ while $K$ is a set of attack spaces}

\textit{Average over attacks. } One simple error function which has been used by prior work is average performance over attacks \citep{TB19} and is defined below.
\begin{equation} \label{eq:avg}
    L_{\text{avg}}(M, K) = \frac{1}{k}\mathbb{E}_{(x, y) \sim \mathcal{D}}\sum_{T \in K} \ell(M(P_T(x, y, M)), y)
\end{equation}

\textit{Worst case over attacks. } Another error function used by prior works \citep{TB19, MainiWK20} measures the worst-case performance over attacks, which can also be thought of as robustness against the union of all attack spaces.  We provide the definition below.
\begin{multline} \label{eq:worst}
    L_{\text{worst}}(M, K)= \mathbb{E}_{(x, y) \sim \mathcal{D}}\max_{T \in K} \ell(M(P_T(x, y, M)), y)
\end{multline}
 
\textit{Compositions of attacks. } Prior works have also looked at achieving compositional robustness \citep{hsiung2022carben, Hsiung2022towards}.  Here, the attacker can apply all perturbations to a single image sequentially, so the defender would like to achieve robustness against the worst-case sequence of attacks.  We provide the definition below where $P_i$ denotes the worst-case attack corresponding to the $i$th attack space contained in $K$.
\begin{multline} \label{eq:comp}
    L_{\text{comp}}(M, K)= \mathbb{E}_{(x, y) \sim \mathcal{D}}\max_{i_1, ..., i_{|K|} \in \{1...|K|\}}  \\\ell(M(P_{i_1}(...(P_{i_k}(x, y, M))), y)
\end{multline}

The choice of error function specifies the goals of the defender.  In general for security critical applications, worst case and composition loss functions may be more realistic as an attacker would likely continue to exploit any vulnerabilities and in application domains such as the image domain, different perturbations can be composed to create a valid input.

\section{Applications for which Resilience is Important} 
\label{sec:applications}

We now provide some applications in which adversarial resilience is important to further motivate designing adaptive defenses with this property.  Some application domains include the image and language domains in which it is difficult to concretely formalize the entire space of perturbations we would like the model to be robust against.  Due to the difficulty of formulating the space of perturbations, it is difficult to design defenses in these domains since new types of adversarial perturbations can be introduced over time.  Some examples of applications include:
\begin{itemize}[leftmargin=*]
    \item \textbf{Social media filtering.} We want to ensure that we can accurately filter out harmful posts even if an adversary tries to use perturbations via images and text in order to avoid filtering. Since it is difficult to anticipate all forms of these perturbations, it is reasonable to expect that an adversary can succeed at avoiding detection, but ideally a good social media site should actively try to refine the system over time such that the same adversarial perturbations cannot be expoited by an adversary for long periods of time.
    \item \textbf{Surveillance.} In some settings, we may want to monitor and identify harmful behavior through surveillance.  It is important that we can accurately identify this behavior even if an adversary intentionally tries to evade detection by using different clothing designs \citep{xu2020adversarial}, but it may be difficult to anticipate all forms of ways to evade detection at deployment.  However, we would like the surveillance system to be resilient such that the system is not consistently fooled by the same types of adversarial perturbations for long periods of time.
    \item \textbf{Face authentication.} For secure face authentication, it is important that small perturbations which might not be modelled via $\ell_p$ bounded attacks (ie. lighting changes, spatial shifts) do not cause an adversary to succeed at impersonating someone else.  However, it is difficult to anticipate all possible perturbations that can be used by an adversary.  If an adversary does succeed at impersonating someone else, we would hope to quickly detect this security breach and also update the model so that the adversary can no longer exploit this vulnerability.
    \item \textbf{Jailbreaking in LLMs and VLMs. }Research in LLMs and VLMs demonstrate that safety guardrails on these models can be jailbroken by modifying the input in order to cause these models to generate harmful content \citep{zhou2024easyjailbreak,wei2024jailbroken,zou2023universal}.  If we are aware that a jailbreak method exists, we should be able to quickly patch the LLM or VLM to be robust to this jailbreak method.
    \item \textbf{Watermarking.} It is important for watermarking that small imperceptible perturbations of different types (ie. $\ell_p$ perturbations, spatial perturbations, small color shifts) are unable to cause detection of the watermark to fail \citep{jiang2024certifiably}.  In the case that someone in the future discovers a perturbation that is able to cause watermark detection to fail, then the watermarking algorithm should be able to be quickly adapted to minimize misuse of this perturbation.
\end{itemize}

\section{Challenges of an Adversarially Resilient Defense and Open Directions}
\label{sec:ars_challenges}

The problem of adversarial resilience considers an adaptive defense which can update the model based on test-time examples.  We now discuss challenges of implementing an adversarially resilient defense and discuss open directions for research.

\textbf{Lack of labels and corresponding clean inputs. }During test-time the defender only has access to the test input which may have been manipulated by the adversary and does not know the corresponding label or the original clean test input.  Many existing defenses in adversarial robustness look at optimizing robust loss \cite{madry2017towards,TB19,croce2020robustbench} or measuring distance in output from the clean input \citep{zhang2019theoretically,yang2020closer, dai2022formulating}, but during test-time these cannot be measured due to no access to labels or corresponding clean inputs.  This challenge raises the question, is there another objective we can optimize without this information that can enhance robustness of the model when an example from a new attack is given to the model?  If not, can we use a detector to detect new attacks as in \citep{stutz2020confidence} and having a human in the loop to label a few of these examples to work with the defense algorithm?  Alternatively, could we leverage generative models \cite{ho2020denoising,rombach2022high} in order to approximate the clean input?  We note that for these techniques to be used in an adversarial resilient defense, we would need to ensure that the detector or generative model is resilient (or robust) as well.

\textbf{Vulnerability to poisoning. }If the attacker knows that the defense is using test examples to update the model, then the attacker can also use poisoning in order to hurt the model's overall performance (causing the defense to fail to meet the criteria for adversarial resilience as well since it would fail to maintain high clean accuracy).  This makes poisoning a challenge for resilience as well and raises the question of how to design a defense that resistant to poisoning and simultaneously can be used to update the model to be robust against evasion attacks.

\textbf{Forgetting of robustness against previous attacks. } If we consider the case where the attacker does not decide to attack the model until time $t$, then an adaptive defense which uses information from clean test examples before time $t$ to update the model may suffer from forgetting initial robustness. %\sophie{if time run experiment and add small table for this + discussion}
This may cause the model to be less robust than the initial model at deployment making it difficult to maintain good robustness against previously seen attacks.  Even if the attacker decides to attack continuously starting from $t=0$, forgetting can occur as well when the attacker switches to using a new attack type.  We can consider new attack types as analogous to different tasks in continual learning, which is a field in which catastrophic forgetting (forgetting of previous tasks) is a large problem \citep{wang2023comprehensive,MCCLOSKEY1989109,kirkpatrick2017overcoming}.  A potential direction for mitigating forgetting is seeing if approaches in continual learning for reducing forgetting can also help with adversarial resilience.

\textbf{Few shot learning of new attacks. } In order for the defended model to quickly recover from new attacks and minimize the ability of the adversary to exploit the same attack repeatedly, it should be able to learn from new attacks with as few examples as possible.  However, existing works in adversarial robustness demonstrate that methods such as training with perturbed examples require many examples in order to achieve good robust performance \citep{carmon2019unlabeled, sehwag2021robust, gowal2021improving, montasser2021adversarially}, which raises questions about the feasibility of achieving resilience.

\textbf{Test-time adaptation. } Given that an adaptive defense is allowed to use examples in test-time in order to improve the defense, one potential starting point for exploration is whether techniques in test-time adaptation (TTA) \citep{wang2020tent, gong2022note} can be adapted to allow for adaptation to new types of adversarial perturbations.  TTA also raises more questions and challenges in the adversarial resilience problem due to the presence of a test-time adversary.  For example, \citet{wu2023uncovering} demonstrated that TTA is vulnerable to poisoning and must be implemented robustly to avoid having this vulnerability exploited by the attacker.

\textbf{Distribution shifts over time. }Just as the space of attacks can change over time, we may also see changes in distribution over time.  Another interesting problem to study is how in-distribution and out-of-distribution examples can be utilized in order to achieve adversarial resilience in conjunction to distributional shifts.

\textbf{Assumptions for feasibility. }Overall, we discussed many challenges for adversarial resilience which are not present in the original robustness problem.  This raises the question, what assumptions do we need to make about the adversary or the defender in order for adversarial resilience to be feasible?

%% file: sections/7CAR.tex
\label{Sec:CAR}
As discussed in Section \ref{sec:ars_challenges}, designing an adversarially resilient defense has a number of challenges, including being able to detect when new attacks occur and update the model to be robust with only a few examples of the attack and limited labels.  In this section, we introduced a simplified but practical setting of adversarial resilience in which the defender discovers the formulation of the attack space over time.  We can model this through an adaptive defender knowledge which is an analog of the adaptive attacker knowledge set but specifies the attack spaces that the defender is aware about.

\begin{definition}[Adaptive Defender Knowledge Set]
The adaptive defender knowledge set  $\mathbf{K}_{\text{defender}}(t)$ is time dependent function which outputs a set of attack spaces $T: X \to 2^X$ representing the set of attack spaces known to the defender at time $t$. At time $t$, the defense algorithm chosen by the defender can incorporate information about attack spaces in $\mathbf{K}_{\text{defender}}(t)$. 
\end{definition}

We now give an updated adversarial game similar to Definition \ref{def:adv_game} to serve as the basis of defining \adaptiverob~(CAR).  We will then discuss how this definition compares to the single attack robustness game (Definition \ref{def:adv_game}) and then discuss its connection with adversarial resilience.

\begin{definition}[Continual Adaptive Robustness] \label{def:multi_adv_game}
\end{definition}
\begin{enumerate}[leftmargin=*]
\itemsep0em
    \item The defender uses information from $\mathbf{K}_{\text{defender}}$ to select a defense algorithm $\mathcal{A}$.  $\mathcal{A}$ outputs a time-dependent model $\mathbf{M}(t)$ (since at each time step the information in $\mathbf{K}_{\text{defender}}$ can vary).
    \item   The attacker then defines a time-dependent set of attack procedures $\tilde{\textbf{P}}(t)$ where each element $\tilde{P}: X \times Y \times \mathcal{H} \to X$. All attacks in $\tilde{\textbf{P}}(t)$ conform to attack spaces in $\mathbf{K}_{\text{attacker}}(t)$. The attacker can define $\tilde{\textbf{P}}(t)$ based on $\mathcal{A}$ \footnote{Here, the extent to which the attacker can use information about the $\mathcal{A}$ dictates whether this is a white-box or black-box attack.  For example, if the attacker can use information about gradient through each model, this would be a white-box attack.}.
    \item Before deployment, the defender obtains a time dependent model $\mathbf{M}(t) = \mathcal{A}(D_{\text{train}}, \{\mathbf{M}(i) | 0 \le i < t\})$ which can use information from previous time steps.
    \item After deployment, for every test data $(x, y) \sim \mathcal{D}$, the defender evaluates the performance of $\mathbf{M}(t)$ on the perturbed input $\tilde{\textbf{P}}(t)$ for each time $t$.  This performance is assessed using a loss function for multiple attacks $L$ which takes a model and set of perturbations as input.
    \item The defender wins if with high probability the following conditions hold for loss function $L$ which takes a model $M$ and set of attack space $K$ as input: 
    \begin{enumerate}
        \item \textit{Maintain good clean accuracy: }At all times $t$, the defender must achieve $$\mathbb{E}_{(x, y) \sim \mathcal{D}} \ell(h_t(x), y) \le \delta_c$$
        \item \textit{Attacker has access to more attacks than the defender:} If at time $t$, $\mathbf{K}_{\text{attacker}}(t)$ contains an attack space that is not covered by $\mathbf{K}_{\text{defender}}(t)$, then the defender must achieve:
        $$L(M_t, \mathbf{K}_{\text{attacker}}(t')) \le \delta_u$$
        \item \textit{Defender recently learned about new attack:} If time $t$ lies within a small value $\Delta t$ after an introduction of a new attack space at time $t'$ into $\mathbf{K}_{\text{defender}}$, then the defender must achieve:
        $$L(M_t, \mathbf{K}_{\text{defender}}(t')) \le \delta_u$$
        \item \textit{All attacks are known by the defender and the defender should have good robustness on this set: }If the previous 2 cases do not hold at time $t$, then the defender must achieve for $\delta_k \le \delta_u$: $$L(M_t, \mathbf{K}_{\text{defender}}(t)) \le \delta_k$$
        \end{enumerate}
\end{enumerate}  

The steps of in the CAR game (\ref{def:multi_adv_game}) generally follow a similar form to the single attack robustness game (Definition \ref{def:adv_game}), where (1) the defender first selects a defense algorithm, (2) the attacker then chooses attack procedures, (3) the defender uses the algorithm to get a robust model, (4) the attacker generates attacks in test-time, and (5) the defender wins if some criteria hold.  There are three main differences from the single attack robustness (SAR) game which we highlight below:
\begin{enumerate}[leftmargin=*]
    \item \textit{Adaptive defender and attacker knowledge sets:} Unlike the SAR game which assumes that both the defender and attacker have knowledge of the same attack space, the CAR problem uses adaptive defender and attacker knowledge sets in order to model the space of perturbations known by the attacker and defender.  This allows the CAR setting to capture differences in knowledge between the attacker and defender ($\mathbf{K}_{\text{attacker}}(t)$ and $\mathbf{K}_{\text{defender}}(t)$).  These sets are also time dependent, which allows CAR to model the discovery of new attack types by both the attacker and defender over time.
    \item \textit{Defender and attacker choose a time dependent model and attack procedures:} Unlike the SAR game where the attacker chooses a single attack procedure and defense algorithm outputs a single robust model, in the CAR game, these aspects are time dependent due to the addition of new attacks into $\mathbf{K}_{\text{attacker}}$ and $\mathbf{K}_{\text{defender}}$ over time.  For the defender, the defense algorithm outputs $\mathbf{M}(t) = \mathcal{A}(D_{\text{train}}, \{\mathbf{M}(i) | 0 \le i < t\})$; we can visualize this as the defense algorithm outputting a sequence of different models based on information from the training data and previous models in the sequence.
    \item \textit{Modified win criterion:} In the SAR game, we were concerned with only 2 quantities, the robust error measured using the attacker's chosen attack procedure and the clean accuracy.  In CAR, maintaining clean accuracy at all times is also a goal (5a in Definition \ref{def:multi_adv_game}), but the robustness criteria is more complex and depends on the time $t$ and relationship between $\mathbf{K}_{\text{attacker}}(t)$ and $\mathbf{K}_{\text{defender}}(t)$.  The defender essentially has 2 thresholds for robust error, $\delta_u \ge \delta_k$. $\delta_u$ is used for attacks unforeseen by the defender (5b in Definition \ref{def:multi_adv_game}) and when the defender has recently (within $\Delta t$ time) become aware of a new possible attack (5c in Definition \ref{def:multi_adv_game}).  If neither of these 2 cases hold, then the defender knows about all attack types in $\mathbf{K}_{\text{attacker}}(t)$ and does not need time to adjust to a new attack, so it must achieve lower robust error $\delta_k$  (5d in Definition \ref{def:multi_adv_game}).  The combination of criteria 5c and 5d ensure that the defender's algorithm is able to output an updated model that is robust to new known attacks within a short $\Delta t$ period of time.
\end{enumerate}

Additionally, we note that SAR game can be considered a special instance of CAR where $\mathbf{K}_{\text{defender}}(t)$ is constant across all $t$, $|\mathbf{K}_{\text{defender}}(0)| = 1$, and $\mathbf{K}_{\text{defender}}=\mathbf{K}_{\text{attacker}}$.

\textbf{Connection to Adversarial Resilience. }The win criteria for the defender in the CAR game align with the goals of adversarial resilience (Definition \ref{def:resilience}).  The main difference with the adversarial resilience problem is the introduction of the defender adaptive knowledge set, allowing the defender to have direct access to the formulations of the different attack spaces over time.  Due to this the defense algorithm can use information solely from the training dataset (and outputted models from previous time steps).  This simplification removes many of the challenges faced by adversarial resilience; for instance, since the defender has access to formulations of different attack spaces, they can generate perturbed examples on the training data for use with the defense algorithm.  Because of this, the issue of limited access to labels and poisoning due to adapting the model based on test-time examples is also removed.

\textbf{Significance. }The simplification from adversarial resilience raises the question: is the setting of CAR applicable at all in the real world.  To demonstrate that CAR is still a reasonable setting for study, consider the following three possible events which can occur at some time $t$ after a company deploys a model that is robust against existing attack spaces: 

\begin{enumerate}
\itemsep0em
    \item \textit{A research group publishes a paper introducing a new attack space. } This setting can be modeled by CAR, where at time $t$ when the research group publishes their paper, $\mathbf{K}_{\text{defender}}(t)$ and $\mathbf{K}_{\text{attacker}}(t)$ both update to include this new attack space.
    \item \textit{An adversary discovers a new vulnerability. } This setting can be modeled by CAR, where $\mathbf{K}_{\text{attacker}}(t)$ is updated to include this new attack space.
    \item \textit{The company's security team discovers a new vulnerability. } This setting can be modeled by CAR, where $\mathbf{K}_{\text{defender}}(t)$ is updated to include this new attack space.
\end{enumerate}
Due to the applicability of CAR to these real world settings, we also argue for more research in this direction.
\begin{froval}
    \begin{cfinding}We advocate for research in continual adaptive robustness, a simplified but practical setting of adversarial resilience.
    \end{cfinding}
\end{froval}

\subsection{Open Directions for Research}
Since CAR is new problem setting introduced in this paper, there have not been any works that directly address this problem.  We now discuss open directions of research for improving CAR.

\textbf{Finetuning-based Defense Algorithms. } One potential direction is looking at how to efficiently finetune robust models to be robust against new attacks.  Of existing works aimed towards achieving good sMAR performance, \citet{croce2022adversarial} proposes a finetuning approach which can be applied to the CAR problem.  However, the focus of this work is on robustness against unions of $\ell_p$ attacks and it is unclear whether it performs well when extended to non-$\ell_p$ attacks. Thus, it is currently unclear what finetuning procedures can work well for general attack types.  We encourage more works on improving finetuning but with a broader scope of attacks.

%\textbf{Continual Learning. }We note that the setting of CAR is also similar to continual learning \citep{wang2023comprehensive}, but rather than adapting to new tasks, we adapt to new attacks.  A few interesting directions of research are understanding the connections between continual learning and CAR; for example, do problems faced in continual learning such as catastrophic forgetting arise in the same manner in CAR, and can algorithms for continual learning be modified to address the CAR problem?

\textbf{Standardized Evaluation. }Another important direction is benchmarking.  One basic evaluation procedure is to consider a fixed sequence of different attacks and measure accuracy on this sequence before and after adapting to each attack.  However, it may be that an algorithm can more easily adapt to one ordering of the attacks than another so having a fixed arbitrary sequence may lead to an unfair comparison.  Additionally, using this evaluation procedure does not take into account the runtime of each model update which is critical to CAR.  We invite more research on creating a standardized benchmark for evaluating CAR, which we believe will also help inspire more research in this field.

%% file: sections/5MAR.tex
In Section \ref{Sec:CAR}, we demonstrated continual adaptive robustness's connection to adversarial resilience and its connection to single attack robustness.  We will now begin this section by demonstrating a strong connection between CAR and the problems of simultaneous multiattack robustness (sMAR) and unforeseen attack robustness (UAR).

\begin{figure}[ht]
    \centering
    \includegraphics[width=\columnwidth ]{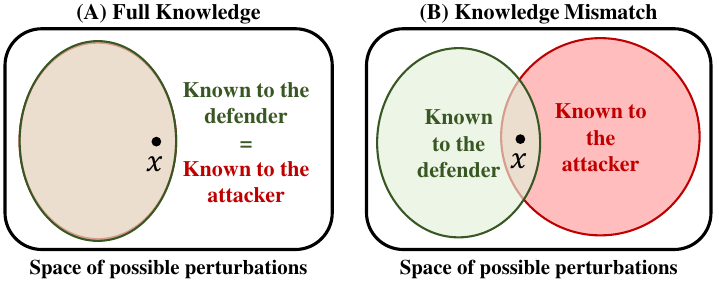}
    \vspace{-15pt}
    \caption{\textbf{Full knowledge vs knowledge mismatch.} The white box represents the space of possible perturbations that we would expect a model to be robust to (ie. space of imperceptible perturbations) which we may not know how to model.  The green oval represents the space of perturbations captured in the defender knowledge set and the red oval represents the space of perturbations captured in the attacker knowledge set. \textbf{(A) Full knowledge} occurs when the defender knows the space known to the attacker while \textbf{(B) Knowledge mismatch} occurs when there exist perturbations known to the attacker that are not known to the defender.  Robustness in this setting corresponds to \textit{unforeseen attack robustness}.}
    \label{fig:fullvpartial}
    \vspace{-10pt}
\end{figure}

\textbf{Connection between CAR to sMAR and UAR. }Similar to single attack robustness, sMAR and UAR can be modelled via the CAR adversarial game (Definition \ref{def:multi_adv_game}).  Specifically, sMAR occurs when $\mathbf{K}_{\text{defender}}(t)$ is constant across all $t$, $|\mathbf{K}_{\text{defender}}(0)| > 1$, and $\mathbf{K}_{\text{defender}}=\mathbf{K}_{\text{attacker}}$. 
 Meanwhile, UAR occurs when $\mathbf{K}_{\text{defender}}(t)$ and $\mathbf{K}_{\text{attacker}}(t)$ are constant across all $t$ and $\bigcup_{T_d \in \mathbf{K}_{\text{defender}}(0)} T_d(x) \subset \bigcup_{T_a \in \mathbf{K}_{\text{attacker}}(0)} T_a(x)$ for some input $x$.  This relation between attacker and defender knowledge sets for UAR is visually depicted in Figure \ref{fig:fullvpartial} and has also been referred to as the setting of \textit{knowledge mismatch} by prior work \citep{dai2023multirobustbench}, while the opposite setting is called the \textit{full knowledge} setting.

 Another way in which we can see that sMAR and UAR are connected to CAR is by the CAR setting at a fixed time $t$.  When there are multiple attacks in $\mathbf{K}_{\text{defender}}(t)$, the defender faces the problem of sMAR; the defender would like gain robustness against all the attacks that they know about.  Meanwhile, when $\mathbf{K}_{\text{attacker}}(t)$ contains an attack type that is not present in $\mathbf{K}_{\text{defender}}(t)$ or before the CAR defense algorithm has finished integrating information from a new attack recently introduced in $\mathbf{K}_{\text{defender}}(t)$, the defender faces the problem of UAR.

 \textbf{sMAR and UAR defenses as baselines for CAR. }  Since sMAR defenses integrate information from all known attacks into the defense algorithm (ie. training with all known attack types), a trivial defense for CAR would be repeatedly applying sMAR every time $\mathbf{K}_{\text{defender}}$ expands to incorporate a new attack type.  However, existing defenses for sMAR have poor performance on unforeseen attacks \citep{laidlaw2020perceptual, dai2023multirobustbench} and are generally inefficient since they involve training a model with all attack types from scratch \citep{TB19,MainiWK20}.  We can consider UAR defenses as a potential CAR defense as well, where the CAR algorithms outputs the UAR defended model for all time steps.  However, current UAR defenses are still unable to achieve reasonable robustness scores when evaluated on a wide variety of attacks \citep{dai2023multirobustbench}.  A combination of sMAR and UAR approaches can be used as a baseline for CAR as well; however, research in sMAR and UAR is very limited, and to the best of our knowledge, no prior works look at combining sMAR with UAR.

 Thus, it is clear that there is a strong relationship between sMAR, UAR, and CAR, and \textit{improvements in sMAR and UAR algorithms can also help with CAR}.  This leads to our final argument:
 \begin{froval}
    \begin{cfinding}\label{cfind: justification} We strongly urge the research community to study the understudied problems of sMAR and UAR as improvements in these directions can also improve CAR (and by extension, adversarial resilience).
    \end{cfinding}
\end{froval}

%One direction in adversarial ML research that can be further explored to improve the practicality of algorithms for adversarial robustness is simultaneous multiattack robustness (sMAR).  sMAR aims to create defenses that can \textit{incorporate knowledge of different attacks during training.}

%\textbf{Significance. }Currently, attacks of many different attack spaces exist\citep{szegedy2013intriguing, brown2017adversarial, XiaoZ0HLS18, kang2019robustness, wasserstein_attacks, LaidlawF19}, and ideally, we would like to achieve robustness against all attacks.  Defenses for sMAR can benefit from knowing about new attacks in order to create defenses that are applicable to a wider variety of attacks than single attack defenses.

%% file: sections/8conclusion.tex
%\sophie{discussion: maybe ouside vision domain: speech, language, malware, software}
Prior work \citep{dai2023multirobustbench} demonstrates existing approaches do not achieve good robustness when tested on a wide variety of attacks.  In this position paper, we argue that the problem of robustness itself is limited as it considers the defended model fixed after deployment and does not provide any guidance on next steps if the attacker were to succeed.  However, in practice, we would want to quickly update the deployed model so that vulnerabilities cannot be continuously exploited by the adversary.  To this end, we introduce the problem of \textit{adversarial resilience}, which encourages defenses that are robust against previously encountered attacks and can be quickly updated to be robust against new attacks.  We then outline the challenges of developing an adversarially resilient defense and introduce a simplified problem setting called continual adaptive robustness.  Both these settings capture the goal of quick adaptation for robustness against new attacks.  We then demonstrate the connection between continual adaptive robustness and existing (but understudied) directions of simultaneous multiattack robustness and unforeseen attack robustness.  Our position is three-fold:
\begin{enumerate}
    \item We encourage researchers to consider developing adaptive defenses with the goal of adversarial resilience.
    \item We advocate for research in continual adaptive robustness as it avoids many complications in adversarial resilience due to adapting on test examples.
    \item We urge more research in simultaneous multiattack and unforeseen attack robustness as improvements in these fields can directly be used to improve continual adaptive robustness.
\end{enumerate}

%% file: sections/appendix.tex
\subsection{Instantiating the Single Attack Game}
\label{app:instantiating}
In Section \ref{Sec:background}, we introduced the adversarial game for single attacks and then discussed defenses proposed in prior works.  We will now demonstrate how the adversarial game is captured in the design of existing defenses.

We can define each defense algorithm $\mathcal{A}$ as the composition of a training algorithm and a test-time procedure: $\mathcal{A} = \mathcal{A}_{\text{test}} \circ \mathcal{A}_{\text{train}}$. $\mathcal{A}_{\text{train}}$ is a learning algorithm which takes the training dataset $D_{\text{train}}$ as an input and outputs a model.  Meanwhile $\mathcal{A}_{\text{test}}$ both takes in a model as input and outputs a model.  Since all defenses can be considered either training time or test-time defenses (or the combination of both), this formulation models all existing defenses.

\textbf{Example 1: Adversarial training \citep{madry2017towards, zhang2019theoretically}. }  To model adversarial training, $\mathcal{A}_{\text{train}}$ is the adversarial training algorithm (ie. PGD adversarial training with $\ell_2$ attacks of radius 0.5) while $\mathcal{A}_{\text{test}}$ is the identity function.  Thus $\mathcal{A}$ just returns the model after adversarial training and this model is evaluated with test-time adversarial examples.  The objective of adversarial training is to achieve low robust error which is captured by the last step of the game.

\textbf{Example 2: Randomized smoothing \citep{cohen2019certified} } Randomized smoothing is a test-time defense which adds gaussian noise to the input.  We can model this by allowing $\mathcal{A}_{\text{train}}$ to be standard training with training examples augmented with Gaussian noise.  Then $\mathcal{A}_{\text{test}}$ can be considered a wrapper function which adds Gaussian noise to the input: $(\mathcal{A}_{\text{test}}(M))(x) = M(x + \epsilon)$ where $\epsilon$ is Gaussian noise.  The objective of randomized smoothing is to achieve low certified error which can be considered an upper bound on the loss in the last step of the game.

\subsection{Prior Work in Simultaneous Multiattack Robustness}
\label{app:sMAR}
Prior work in sMAR generally employ adversarial training based approaches in order to achieve robustness on a group of adversaries.  \citet{TB19} propose AVG and MAX algorithms for training for sMAR.  These algorithms employ adversarial training with a modified objective function that directly optimizes for low average loss across attacks or low worst case loss across attacks (as in Equations \ref{eq:avg} and \ref{eq:worst}).  \citet{madaan2020learning} propose using random sampling to choose attacks to apply for each batch during training and propose using a noise generator and extra data in order to further improve robust performance across attacks.

Other works on multiattack defenses focus on defending on unions of $\ell_p$ balls.  \citet{MainiWK20} propose a modification of PGD-based adversarial training, which considers using worst case attack at the end of a PGD iteration to be the starting point for the next PGD iteration.  \citet{Croce020} prove that a model that is robust against $\ell_\infty$ and $\ell_1$ attacks, is also provably robust against other $\ell_p$ attacks (with robust radius as a function of the $\ell_\infty$ and $\ell_1$ robust radius). \citet{croce2022adversarial} build off of this work and propose initially training with a single $\ell_p$ attack and then using finetuning with $\ell_\infty$ and $\ell_1$ attacks in order to obtain a model that is robust against other $\ell_p$ attacks.  \citet{NEURIPS2022_a627b946} propose using regularization with single step-$\ell_p$ attacks for improving efficiency.

While previously discussed works focus on obtaining high average-case and worst-case robust accuracy, \citet{Hsiung2022towards} proposes a training procedure for optimizing over the worst case sequence of attacks for compositional robustness (objective defined in Equation \ref{eq:comp}).  The authors also propose a corresponding benchmark for evaluating robustness on compositions of attacks \citep{hsiung2022carben}.

\subsection{Open Directions for Research in Simultaneous Multiattack Robustness}
While several works have looked at the problem of sMAR, the application of some of these approaches is limited and we still lack an understanding of how well we can expect algorithms to perform in this setting.  We now outline several possible directions for future research on the topic of sMAR.

\textbf{Formulation of attack spaces.}  Currently, a large challenge for sMAR is that we do not have a good formulation of the space of attacks that we would like to be robust to.  More research in formulating attack spaces can help with the design of more defenses for multiple attacks, especially certified defenses which need a rigorous formulation of the attack space.  Currently, there is only 1 work on a certified defense for unions of $\ell_p$ balls \citep{Croce020}.

\textbf{General defenses that can work with any attack.} Several works in sMAR look at achieving robustness specifically against $\ell_p$ balls and take advantage of using knowledge of the exact process for generating adversarial examples \citep{MainiWK20} or geometry of the attack space \citep{Croce020,croce2022adversarial}.  This prevents these approaches from being applicable to attacks of other attacks.  For general attack spaces, there is little work outside of using basic adversarial training with modified objectives for average-case and worst-case multiattack performance \citep{TB19}.  This is a potential space in which more research can occur.

\textbf{Understanding and balancing tradeoffs between attacks and performance on unperturbed examples.} Previous works demonstrate that we can expect tradeoffs in robust accuracy between attacks \citep{TB19} and with clean accuracy \citep{tsipras2019robustness, stutz2019disentangling}, but outside of tradeoffs between different $\ell_p$ attacks \citep{Croce020}, it is unclear how much tradeoff is expected and how to effectively balance these tradeoffs.  In current literature, there are no works suggesting algorithmic approaches to tuning this tradeoff between different attacks.  We encourage more works to study this problem and understanding these tradeoffs can allow for better multiattack defenses.

\textbf{Efficiently training with multiple attacks.}  As new attacks conforming to different attack spaces are introduced, how we can efficiently train with all attacks becomes an important problem.  Currently the best performing methods require generating all attacks for each example during training, which can be computationally expensive, especially as the number of attacks increases, while more efficient approaches \citep{madaan2020learning, croce2022adversarial, NEURIPS2022_a627b946} generally trade-off robust accuracy and have not been tested outside of $\ell_p$ attacks.  Research on improving the efficiency of multiattack training (or efficiency of generating attacks of different attack spaces) can make multiattack defenses more practical in the long run.

\textbf{Ensembling.} One potential direction that may be useful is ensembling.  Since many prior works focus on achieving robustness against single attacks, a possible direction is exploring how to ensemble singly robust models in order to achieve sMAR.  We encourage more research in ensembling for sMAR as it allows improvements in single attack robustness to carry over to the multiattack setting.

\subsection{Prior Work in Unforeseen Attack Robustness}
\label{app:UAR}
Currently, few works have looked at the problem of UAR.  One line of works looks at using approximations to model the true attack space.  \citet{laidlaw2020perceptual} proposes using adversarial training with adversarial examples generated based on perceptually aligned distance metrics (specifically LPIPS metric \citep{zhang2018lpips}) instead of $\ell_p$ metrics.  They find that training with these examples leads to models robust against imperceptible attachs such as $\ell_p$ attacks, spatial attacks \citep{XiaoZ0HLS18}, and color change attacks \citep{LaidlawF19}. \citet{ghazanfari2023r} extend this work by pointing out vulnerabilities the LPIPS metric and designing a modified distance metric R-LPIPS, which they then use with adversarial training.

Another line of works in defense literature uses regularization to improve UAR.  \citet{dai2022formulating} introduce a regularization term called variation regularization which encourages feature representations between inputs lying within attack spaces in the defender knowledge set to be close in $\ell_2$ distance.  They demonstrate that this regularization greatly improves robustness on unforeseen (imperceptible) attacks and combine the regularization with LPIPS training \citep{laidlaw2020perceptual} in order to obtain state-of-the-art performance on UAR.  \citet{jin2020manifold} propose training with a series of regularization terms including regularizing Hamming distance between activation patterns and $\ell_2$ distance of outputs of noisy inputs.  They demonstrate that without adversarial training, they can obtain robust models against $\ell_p$ and Wasserstein attacks \citep{wasserstein_attacks} (although at the cost of a large tradeoff in clean accuracy).

A related line of works looks at the problem of detecting unforeseen examples and abstaining on those examples. \citet{stutz2020confidence}, which uses regularizes training so that $\ell_p$ attacks with larger norm than used in training have more uniform confidence and then uses confidence based thresholding in order to determine whether to predict or abstain.  Meanwhile, \citet{chen2022revisiting} proposes an algorithm for jointly training a classifier network and detector network in order to robustly classify on examples with small $\ell_p$ perturbations and reject those with large $\ell_p$ perturbations.

In terms of evaluating and UAR, \citet{kang2019robustness} propose a set of 18 different attacks outside of $\ell_p$ attacks and a corresponding metric for aggregating performance across these attacks into a score representing the UAR.
 \citet{dai2022formulating} introduce a set of metrics for measuring average case and worst case performance across a set of 9 attacks (including a subset from \citep{kang2019robustness}) at 20 different attack strengths.  They also provide a leaderboard for CIFAR-10 dataset which allows for ranking and visualizing the performance of existing multiattack and unforeseen attack defenses against this set of attacks.

\subsection{Open Directions for Research in Unforeseen Attack Robustness}
We now discuss limitations of existing works and outline several directions for further research.

\textbf{Perturbation modeling with generative models. } Currently the bulk of research in UAR focuses on improving robustness against the space of imperceptible adversaries; however, for computer vision applications there can be adversarial perturbations that are visible (ie. weather conditions \citep{kang2019robustness}, patch attacks \citep{brown2017adversarial}) which we want models to be robust to.  Rather than modeling the space of attacks using a perceptual distance metric, can we use generative models to learn the space of attacks to generate realistic perturbations for use with adversarial training?  This idea has been explored in \citet{wong2020learning} for modeling specific types of perturbations (such as lighting changes) for single attack robustness, but not for UAR.

\textbf{Theoretical study. }There is little theoretical understanding for when we can expect to generalize to unforeseen attacks.  For regularization based approaches, \citet{dai2022formulating} use an argument based on lipschitzness to demonstrate that regularization may help with robustness against unforeseen attacks, but given an attack that a defense may be robust to, there is little understanding of what other types of attacks the defense can generalize to.  For works that propose using approximations to the true attack space to achieve robustness, there is no theoretical work on why this is expected to succeed.  In what setting can noisy approximations of the true attack space allow us to generalize to the true attack space of interest? Further understanding the conditions in which we expect regularization or perturbation modeling to work can help with inspiring the design of defenses.

\textbf{Managing tradeoff with clean accuracy. }\citet{dai2022formulating} and \citet{jin2020manifold} observe significant tradeoffs with clean accuracy when applying regularization for improving UAR.  For example, On CIFAR-10, \citet{jin2020manifold} achieves only 69.95\% clean accuracy in order to obtain robustness on $\ell_\infty$, $\ell_2$, and Wasserstein attacks, while \citet{dai2022formulating} demonstrates that increasing regularization strength, which improves unforeseen robustness, also decreases clean accuracy. 
 This tradeoff is also present in \citet{laidlaw2020perceptual}'s LPIPS trained model; the best performing across unforeseen attacks on CIFAR-10 only achieves 71.6\% robust accuracy. Thus, another direction is understanding why this tradeoff occurs and how we can reduce it.

\textbf{Efficient evaluation. }To properly assess true UAR it is important to have a large variety of strong attacks to evaluate with which may be computationally expensive.  For example, due to large evaluation times, MultiRobustBench \citet{dai2023multirobustbench} is currently restricted to CIFAR-10, and the only architectures present on the leaderboard are ResNet-18, WRN-28-10, and ResNet-50 which are much smaller than the best performing WRN-70-16 models on the Robustbench leaderboard \citep{croce2020robustbench}. The computational inefficiency faced by \citet{dai2023multirobustbench} is due to the 180 total evaluations, 60 of which are AutoAttack \citep{croce2020reliable} evaluations for $\ell_p$ robustness which are known to give a reliable assessment of $\ell_p$ robustness at the cost of computational inefficency.  \citet{kang2019robustness}, despite proposing 18 different attacks, only evaluates with 8 attacks at 3 attack strengths in order to efficiently evaluate on ImageNet.  Thus improving evaluation efficiency and reliability can also help with benchmark new defenses.